\documentclass[letterpaper, 10 pt, conference]{ieeeconf}

\IEEEoverridecommandlockouts
\usepackage{graphicx}
\usepackage{amsmath}
\usepackage{booktabs}
\usepackage{array}
\usepackage{cite}
\usepackage{url}
\title{\vspace{-0.15in}\LARGE \bf
A Direct Rigid Transmission 2-DoF\\
Wrist Extension for Tendon-Driven Hand\vspace{-0.2em}
}

\author{Yujie Pang$^{1,2}$, Sadman Sakib$^{1}$ and Mohammad Abdullah Al Faruque$^{1}$%
\thanks{$^{1}$The authors are with the Nhu Department of Electrical Engineering and Computer Science,
University of California-Irvine, Irvine, CA, USA.
\texttt{\{yujiep2,ssakib,alfaruqu\}@uci.edu}}%
\thanks{$^{2}$Contributed during participation in the UCInspire program at UCI}%
}

\IEEEaftertitletext{\vspace{-0.5\baselineskip}}

\begin{document}

\maketitle
\thispagestyle{empty}
\pagestyle{empty}

\begin{abstract}
\small
Dexterous manipulation in confined spaces requires local control of hand orientation. Without a wrist, a dexterous hand must obtain this local orientation through coordinated motion of the robot arm, often involving several joints and a more complex end-effector path. We present CRAFT-Wrist, a concentric 2-DoF wrist extension that mounts between a robot arm and the CRAFT Hand without modifying the hand. Two XC430-T240BB-T servos drive sideways and front-back rotation through short rigid transmissions. Our initial hardware prototype actuates both axes under load, achieving a demonstrated workspace of $\pm20^{\circ}$ in  radial--ulnar deviation (left--right) and a front-back range of $+80^{\circ}$ in flexion and $-18^{\circ}$ in extension. We characterize the resulting finger-motor loading across several wrist postures and demonstrate how the wrist supplies local orientation in grasping, nail hammering, and blackboard wiping. Demonstrations and assembly instructions are available at \texttt{https://craft-wrist.github.io/}. 
\end{abstract}

\section{INTRODUCTION}

Dexterous manipulation in confined spaces places demanding requirements on hand orientation. In most arm--hand systems, adjusting hand pose near an obstacle or at the boundary of the arm workspace requires coordinated motion of several large arm joints. A wrist adds a local joint close to the task object, where orientation matters most. Recent systems that integrate a wrist with a dexterous hand report clear task-level gains: faster teleoperation and higher success in a ten-user study \cite{rukav2_2026}, reduced task time and arm travel on assembly tasks \cite{softhandmodelw_2026}, and reachability gains in constrained spaces \cite{dexwrist_2025}. These results motivate a modular wrist that can add local orientation to an existing tendon-driven hand while preserving the hand mechanism and control stack.

We present CRAFT-Wrist, a concentric 2-DoF wrist that installs between a robot arm and the tendon-driven CRAFT Hand \cite{craft_hand_2026} without modifying the hand. We report actuation traces of an assembled prototype driven over a demonstrated workspace of $\pm20^{\circ}$ in radial-ulnar deviation (left--right) and a front-back range of $+80^{\circ}$ in flexion and $-18^{\circ}$ in extension, and document three wrist-dominant task demonstrations: grasping, nail hammering, and blackboard wiping. 

% The demonstrations focus on how local wrist motion can replace part of the multi-joint arm repositioning otherwise required to establish the same hand or tool orientation.

\section{RELATED WORK}

Recent systems demonstrate several approaches to augmenting dexterous manipulation with local wrist motion. Ruka-v2 integrates a decoupled parallel wrist with a tendon-driven hand \cite{rukav2_2026}, SoftHand Model-W combines a serial wrist with tendon routing \cite{softhandmodelw_2026}, and DexWrist couples a low-impedance parallel mechanism with quasi-direct-drive actuation \cite{dexwrist_2025}. The broader artificial-wrist design space is surveyed in \cite{bajaj2019wrists}. Table~\ref{tab:systems} compares these systems with CRAFT-Wrist.

CRAFT-Wrist's two compact servos and short rigid transmissions contribute to an axial depth of $79.75$~mm, which is shorter than Ruka-v2's and DexWrist's reported depth. Unlike DexWrist, CRAFT-Wrist does not require custom planetary gearboxes or separate motor controllers. While CRAFT-Wrist and Ruka-v2 both use commercial servos, the two XM430-W210-R wrist motors in Ruka-v2 cost $2.42\times$ that of CRAFT-Wrist's two XC430-T240BB-T motors. Therefore, CRAFT-Wrist design prioritizes reduced axial packaging, simple integration, lower cost, and short, rigid load paths.

\begin{table*}[t]
\centering
\caption{Comparison between CRAFT-Wrist and three other 2-DoF wrists.}
\label{tab:systems}
\footnotesize
\setlength{\tabcolsep}{3pt}
\renewcommand{\arraystretch}{1.05}
\begin{tabular*}{\textwidth}{@{\extracolsep{\fill}}lcccccc@{}}
\toprule
System & Mechanism & Wrist drive & Co-centered? & F/E range & R/U range & Axial depth \\
\midrule
SoftHand Model-W \cite{softhandmodelw_2026} & Serial RR & Tendon & No; 34\,mm offset & $\pm90^{\circ}$ & $\pm30^{\circ}$ & 41\,mm \\
Ruka-v2 \cite{rukav2_2026} & Spherical parallel & Rigid linkage & Yes & $-30/{+}45^{\circ}$ & $\pm35^{\circ}$ & 106.0\,mm \\
DexWrist \cite{dexwrist_2025} & Spherical parallel & 13:1 custom quasi-direct-drive & Yes & $\pm40^{\circ}$ & $\pm40^{\circ}$ & 178.2\,mm \\
CRAFT-Wrist & Spherical parallel & Short rigid transmission & Yes & $-18/{+}80^{\circ}$ & $\pm20^{\circ}$ & 79.75\,mm \\
\bottomrule
\end{tabular*}\\[1pt]
% {\scriptsize $^{*}$CAD-derived neutral-pose axial depth; other dimensions are reported by the cited sources.}
\end{table*}

\section{SYSTEM DESIGN}

\begin{figure}[t]
\centering
\includegraphics[width=\columnwidth,keepaspectratio]{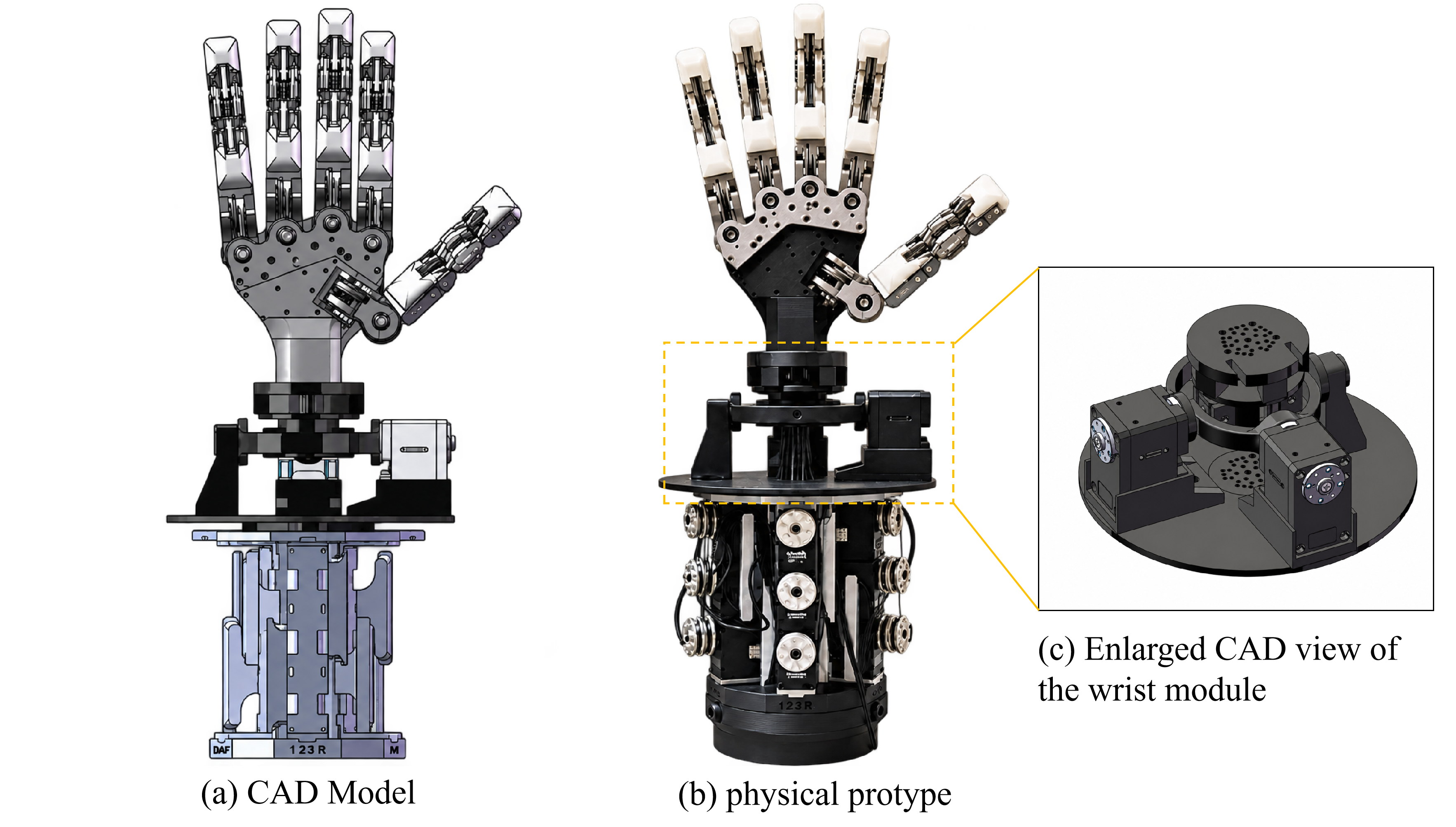}
\caption{CRAFT-Wrist platform: CAD assembly (left), physical prototype (center), and enlarged wrist-module CAD view (right).}
\label{fig:installed}
\end{figure}

CRAFT-Wrist provides two orthogonal axes for radial--ulnar deviation and front--back motion, actuated by two XC430-T240BB-T servos arranged around a compact central wrist structure (Fig.~\ref{fig:installed}). Each axis is driven through a short rigid transmission, giving a direct, deterministic load path from servo to joint that is largely insensitive to tendon pretension, friction, elongation, and routing changes---an advantage over the long tendon runs used in distal wrist designs. The two axes share a common wrist center rather than being stacked in a serial chain. This follows the wrist-center concept of Ruka-v2 \cite{rukav2_2026} while carrying the larger distal mass of the CRAFT Hand without redesigning it. The wrist sits on an independent load-bearing frame between arm and hand and connects to the hand through a defined interface, leaving the hand's mechanics, electronics, and finger-control stack unmodified. Wear-prone parts are detachable and can be replaced simply, easing maintenance without redesigning the hand. The assembled platform therefore adds local hand orientation while preserving the hand's original mechanics and finger-control stack.

Both wrist joints are driven by XC430-T240BB-T DYNAMIXEL servos in position-control mode with configured torque limits, the same actuation family as the hand's finger stack. Each axis is commanded through a short rigid transmission, so the servo command-to-joint relationship is direct and repeatable. 
% The assembled prototype demonstrates a radial--ulnar deviation range of $\pm20^{\circ}$ and a front--back range of $-18^{\circ}$ in extension and $+80^{\circ}$ in flexion under load.

\section{WRIST--FINGER COUPLING CHARACTERIZATION}
\label{sec:motor_load}

To examine whether wrist posture affects finger actuation, we fixed the wrist at
$15^{\circ}$, $30^{\circ}$, $45^{\circ}$, and $60^{\circ}$ and commanded finger
motor~2 through the same outward-and-return sweep. The finger reached the
prescribed travel at every tested wrist posture.

Nevertheless, the motor-current traces changed with wrist posture and sweep
direction: return loading was similar at $15^{\circ}$ and $30^{\circ}$,
increased at $45^{\circ}$, and peaked at $60^{\circ}$ ($300~\mathrm{mA}$,
about 17\% of the motor's $1750~\mathrm{mA}$ limit). Thus, the observed wrist--finger coupling did not
prevent the finger from reaching its commanded positions or cause an
over-current event. The separation between outward and return traces is consistent with
direction-dependent effects such as tendon friction, routing hysteresis, or
stick--slip motion.

% Because each wrist posture was tested with one
% outward-and-return sweep and motor current was not calibrated to output torque,
% these measurements are treated as a preliminary characterization of
% wrist--finger coupling rather than a complete mechanical identification.

\section{WRIST-DOMINANT TASK DEMONSTRATIONS}

We mounted the wrist prototype with a CRAFT hand on a 6-DOF Dobot CR3 arm, and evaluated it in three actions where local wrist orientation directly shapes contact between the hand, tool, and environment. Without a wrist, the same orientation must instead be generated upstream through coordinated shoulder, elbow, and forearm motion. This can require larger joint excursions and repeated repositioning of the arm near obstacles or the edge of the arm workspace.

% This can require larger joint excursions, a curved end-effector path, and repeated repositioning of the arm, particularly near obstacles or the edge of the arm workspace.

\begin{figure}[t]
\centering
\includegraphics[width=\columnwidth,keepaspectratio]{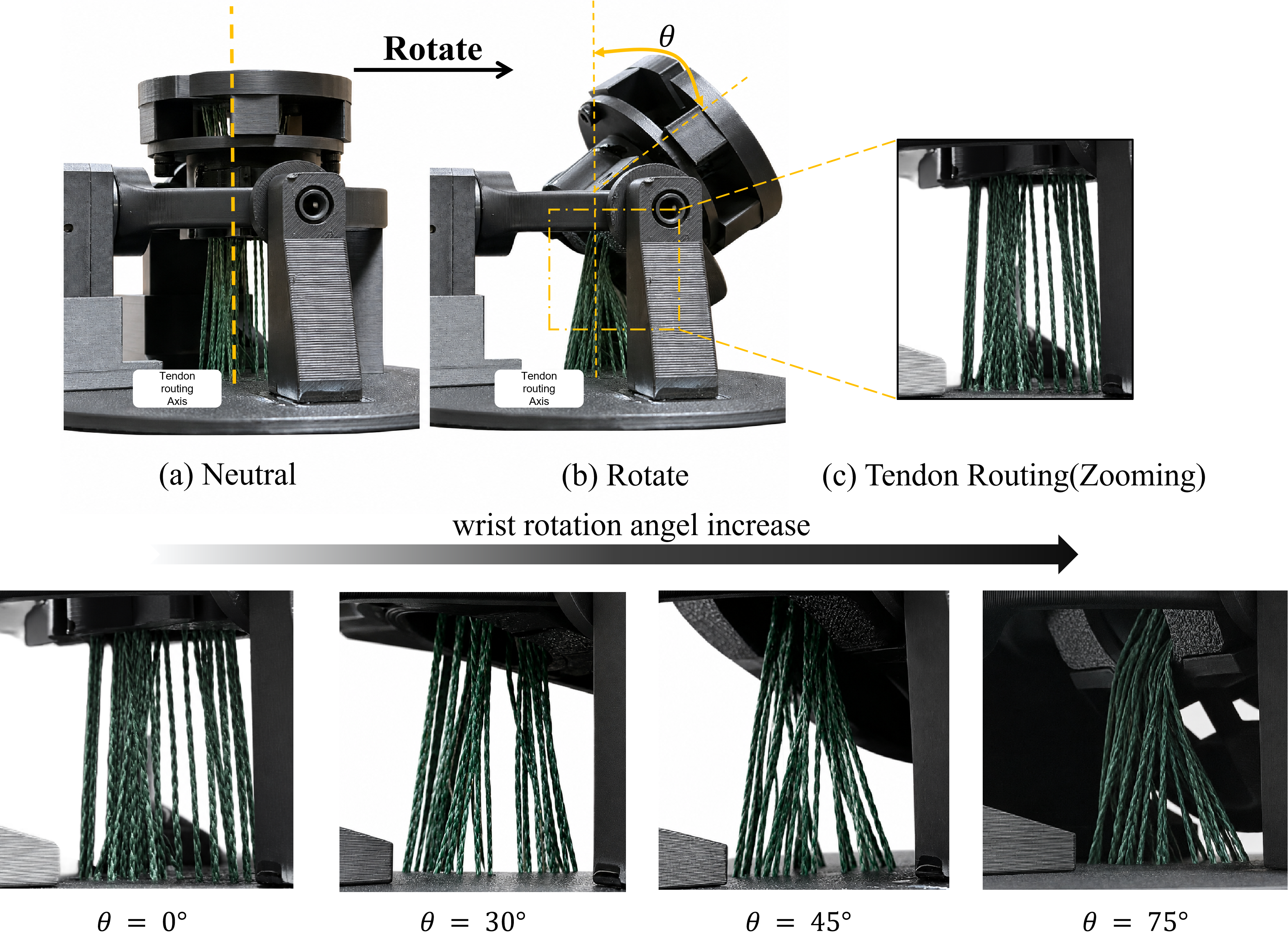}
\caption{Available wrist rotation and tendon routing. The neutral and rotated poses show the local orientation change, while the lower images provide closer views at increasing wrist flexion angles.}
\label{fig:coupling}
\end{figure}

\textit{Grasping and reorientation:} The hand approaches an object, rotates the wrist to align the palm and fingers with the grasp direction, closes the fingers, and reorients while maintaining the grasp. Without a wrist, alignment and reorientation would instead require a coordinated change in the arm pose, which can disturb the approach path or translate the grasped object.

% The wrist can correct the final approach angle.  during reorientation

\textit{Nail hammering:} The hand first grasps the hammer. The wrist then aligns the head with the nail axis and restores alignment between strikes. Without a wrist, the same correction couples hammer-head orientation to elbow and forearm motion, making each recovery more complex.

\textit{Blackboard wiping:} The hand holds an eraser while the wrist sets and adjusts its contact angle during planar wiping strokes, letting the arm supply the broad translation. A wristless hand would instead need continuous arm reorientation to follow the wiping path.

% Together, these tasks show the wrist acting as a local orientation stage between the arm's gross positioning and the hand's contact action.

% The present paper describes this task-level role without claiming a measured reduction in arm travel; a direct comparison of joint excursion, end-effector path complexity, and task performance with and without the wrist is left for future work.

\section{CONCLUSION}
CRAFT-Wrist provides a 2-DoF wrist for tendon-driven hands and adds local
orientation without changing the hand mechanism or finger-control stack.
Wrist posture and motion direction affect finger-motor loading, but the
finger still reaches its commanded position in every tested configuration,
with peak current ($300~\mathrm{mA}$) well under the motor's
$1750~\mathrm{mA}$ limit. The grasping, nail-hammering, and
blackboard-wiping demonstrations show how the wrist can handle local
alignment while the arm supplies gross positioning, avoiding the more
complex coordinated arm motion a wristless hand would require. A repeated
load study and quantitative comparison of arm motion are left for future
work.

{\scriptsize
\enlargethispage{4\baselineskip}
\bibliographystyle{IEEEtran}
\bibliography{references}
}

\end{document}